\documentclass[letterpaper, 10 pt, conference]{ieeeconf}  
\usepackage[T1]{fontenc}

\IEEEoverridecommandlockouts                              

\usepackage{amsmath} 
\usepackage{graphicx} 
\usepackage{cite} 
\usepackage{booktabs}
\usepackage{hyperref}
\usepackage{tabularx}
\usepackage{listings}
\usepackage{xcolor} 
\usepackage{multirow} 

\title{\LARGE \bf
Reinforcement Learning with Curriculum-inspired Adaptive Direct Policy Guidance for Truck Dispatching
}

\author{Shi Meng$^{1}$, Bin Tian$^{2,*}$, Xiaotong Zhang$^{3}$
\thanks{*This work was supported by the Key-Area Research and Development Program of Guangdong Province (2020B0909050001), the Natural Science Foundation of Hebei Province (2021402011), the National Key Research and Development Program of China (2022YFB4703700).}
\thanks{$^{1}$Shi Meng is with the National Key Laboratory for Multi-modal Artificial Intelligence Systems, Institute of Automation, Chinese Academy of Sciences, and with the School of Artificial Intelligence, University of Chinese Academy of Sciences, Beijing 100190, China
        {\tt\small mengshi2022@ia.ac.cn}}%
\thanks{$^{2}$Bin Tian is with the National Key Laboratory for Multi-modal Artificial Intelligence Systems, Institute of Automation, Chinese Academy of Sciences, and with the School of Artificial Intelligence, University of Chinese Academy of Sciences, Beijing 100190, China
        {\tt\small bin.tian@ia.ac.cn}}%
\thanks{$^{3}$X. Zhang is with the National Key Laboratory for Multi-modal Artificial Intelligence Systems, Institute of Automation, Chinese Academy of Sciences, and with the School of Artificial Intelligence, University of Chinese Academy of Sciences, Beijing 100190, China
        {\tt\small zhangxiaotong2023@ia.ac.cn}}%
\thanks{Corresponding author: Bin Tian}%
        }

\begin{document}

\maketitle
\thispagestyle{empty}
\pagestyle{empty}

\begin{abstract}
Efficient truck dispatching via Reinforcement Learning (RL) in open-pit mining is often hindered by reliance on complex reward engineering and value-based methods. This paper introduces \textbf{Curriculum-inspired Adaptive Direct Policy Guidance}, a novel curriculum learning strategy for policy-based RL to address these issues. We adapt Proximal Policy Optimization (PPO) for mine dispatching's uneven decision intervals using time deltas in Temporal Difference and Generalized Advantage Estimation, and employ a Shortest Processing Time teacher policy for guided exploration via policy regularization and adaptive guidance.  Evaluations in OpenMines demonstrate our approach yields a \textbf{10\% performance gain} and faster convergence over standard PPO across sparse and dense reward settings, showcasing improved robustness to reward design.  This direct policy guidance method provides a general and effective curriculum learning technique for RL-based truck dispatching, enabling future work on advanced architectures. 
\end{abstract}

\section{INTRODUCTION}
Efficient management of mine fleet transportation systems is paramount for maximizing productivity in open-pit mining operations~\cite{moradi_afrapoli_mining_2019}.  Fueled by advancements in artificial intelligence and autonomous driving, recent scholarly efforts have increasingly focused on achieving unmanned and highly efficient mining systems. The efficacy of these systems hinges on the performance of truck dispatching algorithms, which are evaluated by key metrics such as shift production tonnage and match factor. To enhance the adaptability and operational performance of these algorithms, diverse approaches have been explored, including optimization and meta-heuristic methods~\cite{mohtasham_optimization_2021,moradi_afrapoli_multiple_2019,both_joint_2020,zhang_scheduling_2021,zhang_real-time_2022}, reinforcement learning (RL) techniques~\cite{noriega2025deep,huo_reinforcement_2023,zhang2023vehicle,zhang_dynamic_2020}, and hybrid dispatching algorithms integrating large models with conventional scheduling~\cite{chen2024llm}.

During operation, trucks in a mine fleet navigate through several interconnected subsystems. In typical open-pit mines, haulage is structured in three eight-hour shifts. Each shift commences with trucks departing from a charging site, initiating a dispatch request for a loading site assignment—termed an "init order".  Loading sites, often spatially clustered, require trucks to queue for service from shovels, which may be heterogeneous. Once loaded, trucks request a "haul order" to the most suitable dumping site, such as a waste dump or crusher, where multiple unloading positions are available. Queuing at dumping sites is generally limited to scenarios with high traffic to a specific site. Upon unloading, trucks request a "load order" to return to an optimal loading site. Throughout a shift, trucks continuously cycle through states of waiting for loading, loading, hauling loaded material, waiting to unload, unloading, and hauling empty. In automated mines, trajectory planning for multi-truck systems within loading and dumping zones is critical to minimize non-productive time. On haul roads, speed limits (e.g., 25 km/h) and no-overtaking policies are enforced for safety, with signal lights at key intersections to balance safety and throughput. The inherent nature of open-pit mining introduces unplanned stochasticity, with equipment requiring periodic maintenance, shovels potentially awaiting blasting, and roads subject to temporary closures for maintenance. These factors present significant challenges to traditional, optimization-based dispatching methods, which struggle with such dynamic uncertainties.

Currently, fixed shovel-truck strategies dominate in practical open-pit mine operations. This experience-based heuristic, prevalent in less automated mines, proportionally allocates truck capacity based on shovel production rates, with trucks dedicated to specific shovel-dump site pairs. Simpler strategies based on shortest processing time~\cite{rose_shortest_2001} and shortest queue length~\cite{subtil_practical_2011} are also utilized.

Optimization-based dispatching algorithms often formulate the problem as a mixed-integer optimization, constrained by loading/dumping capacities, road distances, vehicle speeds, and payloads~\cite{moradi_afrapoli_mining_2019}. Weighted objective functions, considering cost, emissions, and production, are solved using mixed-integer programming solvers (e.g., Gurobi, CPLEX), often followed by meta-heuristic post-optimization. While effective in static environments, these pre-calculated schedules lose optimality when faced with real-world stochastic events. Recalculating optimal schedules in response to every parameter change becomes computationally intensive and may not yield truly optimal solutions over extended operations.

In contrast, reinforcement learning-based dispatching algorithms~\cite{huo_reinforcement_2023,zhang2023vehicle} frame truck order assignment as a sequential decision-making problem. By iteratively computing truck orders in response to system states, RL inherently addresses uncertainty and dynamic events through learned policies, rather than explicitly solving for a static schedule. This approach trains an optimal policy via custom reward functions, offering robustness to unforeseen disruptions. However, mine dispatching presents a complex RL environment characterized by uneven decision intervals and sparse, delayed rewards, unlike typical Gym or Isaac Gym control environments with equidistant decision-making. Traditional temporal difference RL algorithms, which assume uniform decision intervals, face challenges in value function learning within this context. Furthermore, the sparsity and delay in environmental rewards often necessitate manual reward shaping and teacher policies to overcome cold-start issues, adding complexity to the training process.

\section{Related Work}
Reinforcement learning methodologies are broadly categorized into policy-based and value-based approaches. Proximal Policy Optimization (PPO)~\cite{schulman2017proximal}, a prominent policy-based algorithm, effectively balances exploration and stability by constraining policy updates. Its efficacy in complex domains is well-documented, exemplified by OpenAI Five~\cite{berner2019dota}, which used PPO to achieve superhuman performance in Dota 2, and its adoption as a baseline in complex scenarios like Reinforcement Learning from Human Feedback (RLHF)~\cite{ouyang2022training}. Deep Q-Networks (DQN)~\cite{mnih2013playing}, a foundational value-based algorithm introduced for Atari games, has also been adapted for mine dispatching. Dueling DQN~\cite{wang2016dueling}, an enhancement to DQN incorporating an action advantage function, improves convergence and stability, making it a frequently used tool in mine dispatching RL applications~\cite{hazrathosseini2024transition}.

Current RL solutions for mine truck dispatching predominantly favor value-based methods, with policy-based explorations remaining limited~\cite{hazrathosseini2024transition}. Zhang et al.~\cite{zhang2023vehicle} addressed mine fleet dispatching by designing immediate and true value rewards based on service time, employing exponential functions for reward smoothing and simulating shovel availability. They introduced a prioritized experience replay mechanism for Dueling DQN to implement shortest-service-time-based curriculum learning, evaluating performance across varying fleet sizes. However, their Momentum Boosting Strategy (MBS) is intrinsically value-based and not transferable to policy-based algorithms. Similarly, Noriega et al.~\cite{noriega2025deep} utilized DDQN with a fine-grained reward function aimed at achieving target production, using hierarchical rewards and penalties to mitigate cold-start issues, validated in a mine simulation. Huo et al.~\cite{huo_reinforcement_2023} developed an RL-based simulation environment and Q-learning approach, modeling vehicle emissions and comparing performance against rule-based methods for dispatch error and emissions, although focusing only on homogenous trucks with limited training details disclosed.

In summary, existing RL-based dispatching research grapples with: 1) a significant reliance on reward function engineering and shaping to address cold-start problems; 2) a relative scarcity of policy-based approaches despite their broader success in other fields; 3) a lack of versatile curriculum learning methodologies. To overcome these challenges, this paper introduces a general curriculum training strategy based on direct policy guidance, evaluated using policy-based PPO. Experimental results demonstrate that this strategy substantially enhances agent convergence speed, exhibits robustness to reward function design, and is broadly compatible with diverse RL algorithms.

\section{METHOD}
\subsection{Problem Formulation}
From a reinforcement learning perspective, we have formulated the truck dispatching algorithm problem of open-pit mines, referencing previous work, and encapsulated OpenMines using the gym API.



    

\subsubsection{State Space}
To better satisfy the Markov property and enhance model performance, we incorporated extensive environmental information. The observation space for the dispatching agent is primarily divided into Order State, Truck Self State, Road States, and Target States. Relevant processing information and methods are detailed in Table \ref{tab:state_space}. Here, M, N, and K represent the number of loading sites, dumping sites, and trucks, respectively. In this experiment, M=5, N=5, and K=71.

\begin{table*}[t]
\caption{State Space Definition}
\label{tab:state_space}
\centering
\scriptsize
\begin{tabular}{|p{2.2cm}|p{2.8cm}|p{8.5cm}|c|}
\hline
\textbf{Feature Group} & \textbf{Feature Name} & \textbf{Description} & \textbf{Dim.} \\
\hline

\multirow{4}{*}{Order State}
 & \texttt{event\_type}
   & One-hot indicating \{init, haul, load\}
   & 3 \\ \cline{2-4}
 & \texttt{time\_delta}
   & Elapsed time since last dispatch
   & 1 \\ \cline{2-4}
 & \texttt{time\_now}
   & Normalized current simulation time
   & 1 \\ \cline{2-4}
 & \texttt{time\_left}
   & Remaining ratio (1 $-$ time\_now)
   & 1 \\
\hline

\multirow{2}{*}{Truck Self State}
 & \texttt{truck\_location\_onehot}
   & Current location of the truck (One-hot)
   & M+N+1 \\ \cline{2-4}
 & \texttt{truck\_features}
   & \{\,log(load+1), log(cycle\_time+1)\}
   & 2 \\
\hline

\multirow{4}{*}{Road States}
 & \texttt{travel\_time}
   & Estimated travel time based on distance/speed for three routes
   & max(M,N) \\ \cline{2-4}
 & \texttt{truck\_counts}
   & Number of trucks on each road (normalized)
   & max(M,N) \\ \cline{2-4}
 & \texttt{road\_dist}
   & Road distance information for the three routes
   & max(M,N) \\ \cline{2-4}
 & \texttt{road\_jam}
   & Traffic jam counts on each road
   & max(M,N) \\
\hline

\multirow{6}{*}{Target States}
 & \texttt{est\_wait}
   & log of estimated waiting time (including en-route trucks)
   & max(M, N) \\ \cline{2-4}
 & \texttt{tar\_wait\_time}
   & log of queue waiting time (excluding en-route)
   & max(M, N) \\ \cline{2-4}
 & \texttt{queue\_lens}
   & Queue length at each target (normalized)
   & max(M, N) \\ \cline{2-4}
 & \texttt{tar\_capa}
   & log of target capacity
   & max(M, N) \\ \cline{2-4}
 & \texttt{ability\_ratio}
   & Service ratio (reflecting maintenance state)
   & max(M, N) \\ \cline{2-4}
 & \texttt{produced\_tons}
   & log of the ore tonnage already produced
   & max(M, N) \\
\hline

\end{tabular}
\end{table*}

\subsubsection{Action Space}
The action space for the reinforcement learning agent is defined as the set of target loading/dumping sites that a truck, currently requesting dispatch, can proceed to from its current road network location. When a vehicle is at the parking site (ChargingSite), it can only proceed to a loading site. When at a loading site, it can only proceed to a dumping site. When at a dumping site, it can only proceed to a loading site. This design avoids errors caused by agent dispatching trucks to ambiguous locations~\cite{huo_reinforcement_2023}.

\subsubsection{Reward Design}
Reward function design is a significant challenge in reinforcement learning training. Mine dispatching scenarios typically use centralized policies to train agents, with states and actions of multiple trucks intermixed at hundreds of levels. If mine production is solely used as a reward, it leads to problems of reward sparsity and difficulty in attribution. Therefore, most studies employ carefully tuned dense rewards to guide agents to learn in a specific direction~\cite{noriega2025deep,huo_reinforcement_2023,zhang_dynamic_2020,zhang2023vehicle}. Here, we define two types of rewards, dense and sparse, to validate the effectiveness of our direct policy guidance reinforcement learning, as detailed in Table~\ref{tab:reward_function}. For the sparse reward setting, only `final\_tons\_reward` is used, with coefficients for other components set to 0.

\begin{table*}[t]
\caption{Reward Function Design}
\label{tab:reward_function}
\centering
\scriptsize
\begin{tabular}{|p{2.8cm}|p{6.5cm}|c|}
\hline
\textbf{Reward Component} & \textbf{Description} & \textbf{Reward Coefficient (Sparse / Dense)} \\
\hline
\texttt{final\_tons\_reward} & Incentivizes total mine output at episode end. & $0.1 \times \text{final\_tons}$ / $0.1 \times \text{final\_tons}$ \\
\hline
\texttt{delta\_tons\_reward} & Encourages incremental production during simulation. & $0$ / $2.0 \times \log(\text{delta\_tons} + 1)$ \\
\hline
\texttt{wait\_penalty} & Penalizes truck wait time. & $0$ / $-0.5 \times \text{wait\_duration}$ \\
\hline
\texttt{service\_penalty} & Penalizes time spent at loading/dumping sites. & $0$ / $-0.1 \times \text{service\_duration}$ \\
\hline
\texttt{jam\_penalty} & Penalizes time lost due to traffic jams. & $0$ / $-0.1 \times \text{jam\_duration}$ \\
\hline
\texttt{move\_penalty} & Penalizes excessive truck travel time. & $0$ / $-0.01 \times \text{move\_duration}$ \\
\hline
\end{tabular}
\end{table*}

\subsection{Reinforcement Learning with Adaptive Direct Policy Guidance}

\subsubsection{Algorithm Baseline Implementation}
To implement the standard PPO algorithm, our algorithm implementation references the work of John Schulman et al.~\cite{schulman2017proximal} and Huang Shengyi et al.~\cite{huang2022cleanrl}, adjusted for the unique characteristics of mine fleet dispatching scenarios and general reinforcement learning scenarios. In mine fleet dispatching scenarios, multiple trucks typically request dispatching from the same agent. Each request brings a state $s$, and the agent's decision is considered an action $a$. However, the time span between agent decisions is uneven in mine scenarios, unlike traditional game and control simulation scenarios where it is usually uniform. After calculating a Rollout, the PPO algorithm learns new and old policies in epochs, during which both the Policy network and Value network are updated. Since the Value network typically uses the Temporal Difference (TD) method for learning, as shown in Equation~\ref{eq:td-deltat}, uneven time spans can cause difficulties for value function and GAE learning. To maintain the uniformity of temporal reward discounting over time, we used $\Delta t$ (time since the last dispatch) as a power regularization coefficient, allowing the time discount to better match the actual elapsed time. Similar modifications are involved in the Generalized Advantage Estimation (GAE) calculation process, as shown in Equation~\ref{eq:gae-deltat}.

In the case of a decision interval of $\Delta t$, the Temporal Difference (TD) target can be expressed as:
\begin{equation}\label{eq:td-deltat}
V(s_t) \;=\; r_t \;+\; \gamma^{(\Delta t)} \, V(s_{t+1})
\end{equation}
where $r_t$ is the immediate reward or return obtained from transitioning from state $s_t$ to $s_{t+1}$, and $\gamma^{(\Delta t)}$ is the discount factor modified based on $\Delta t$.

When using Generalized Advantage Estimation (GAE), the calculation formula for the advantage function $A_t$ in uneven time-step scenarios can be written as Equation~\ref{eq:gae-deltat} and Equation~\ref{eq:delta_t}:
\begin{align}
\delta_{t}
&=
r_t \;+\;\gamma^{(\Delta t_t)}\,V(s_{t+1}) \;-\; V(s_t),
\label{eq:delta_t} \\[6pt]
A_t^{\text{GAE}(\gamma,\lambda)}
&=
\sum_{l=0}^{\infty}\;
\Bigl[\bigl(\gamma^{(\Delta t_{t+l})}\,\lambda^{(\Delta t_{t+l})}\bigr)^l
\,\delta_{t+l}\Bigr].
\label{eq:gae-deltat}
\end{align}
where $\lambda\in(0,1)$ is the trace decay coefficient of GAE, also with a time-step modification of $\lambda^{(\Delta t)}=\lambda^{\Delta t}$. At this point, each $\delta_{t}$ is individually discounted and decayed according to its corresponding interval $\Delta t_t$, thereby more accurately measuring the "value difference after $\Delta t_t$ time has elapsed".
Through this processing method, traditional RL algorithms can maintain training stability in non-uniform time interval scenarios, making value estimation more closely aligned with the passage of time in the real environment.

\subsubsection{Teacher Policy Term}
The teacher policy is responsible for balancing teaching and exploration in the early stages of training, avoiding difficulties in initial training convergence. We introduce the existing supervised Shortest Processing Time strategy~\cite{meng_openmines_2024} to construct a Loss regularization term as shown in Equation~\ref{eq:guide_loss}.

\begin{equation}\label{eq:guide_loss}
{\text{GuideLoss}} \;=\; \frac{1}{N}\sum_{i=1}^{N}
\log\pi_{\theta}(a_i^{sug}|s_i)
\end{equation}

Where N represents the total number of samples in a minibatch, and $a_i^{sug}$ represents the suggested action of the teacher policy in state $s_i$.

\subsubsection{Adaptive Policy Guidance Coefficient}
We set an adaptive policy guidance coefficient $C_{teacher}$ to characterize the current policy's learning acceptance of the teacher policy, as referenced in Equation~\ref{eq:c_teacher}. A higher acceptance indicates that the current policy is closer to the teacher policy. When the model, under the guidance of the teacher policy, exceeds a preset baseline performance (e.g., production tonnage), teaching of the policy in training is stopped, and $guide\_coef$ is set to 0.

\begin{equation}
c_{\text{teacher}} \;=\;
  \frac{1}{N}\sum_{i=1}^{N}
  \exp\!\bigl(\log \,\pi_\theta\!\bigl(a_i^\text{sug}\mid s_i\bigr)\bigr)
\label{eq:c_teacher}
\end{equation}

\begin{align}
\text{guide\_coef}
&=
\begin{cases}
0,
& \text{if } \text{tons} \;\ge\; \text{base\_tons},\\[6pt]
\alpha\,\bigl(1 - c_{\text{teacher}}\bigr),
& \text{otherwise}.
\end{cases}
\label{eq:guide_coef}
\end{align}
where `tons' refers to the current ore tonnage produced, and `base\_tons' is the pre-set baseline production tonnage.

\section{EXPERIMENT}
\subsection{Environment Setup}
We employ the OpenMines~\cite{meng_openmines_2024} mine simulation environment. For the mine environment configuration, we referenced mine operation information from the Huolinhe North Open-Pit Coal Mine in September 2022, desensitizing relevant data for the OpenMines environment setup. In experiments, we used a time unit of 1 minute and trained for 4 hours with three different fleets totaling 71 trucks, 21 heterogeneous shovels, five loading sites, and five dumping sites. Unlike OpenMines~\cite{meng_openmines_2024} and Xiaotong Zhang et al.~\cite{zhang2023vehicle} who used equal round-trip distances in their road network abstraction, we differentiated the round-trip distances between loading and unloading points. This differentiation altered the performance of some strategies in OpenMines (such as FixedGroupStrategy) due to its greedy strategy in unloading point selection.

\subsection{Training}
\begin{figure}[t]
  \centering
   \includegraphics[width=0.5\textwidth]{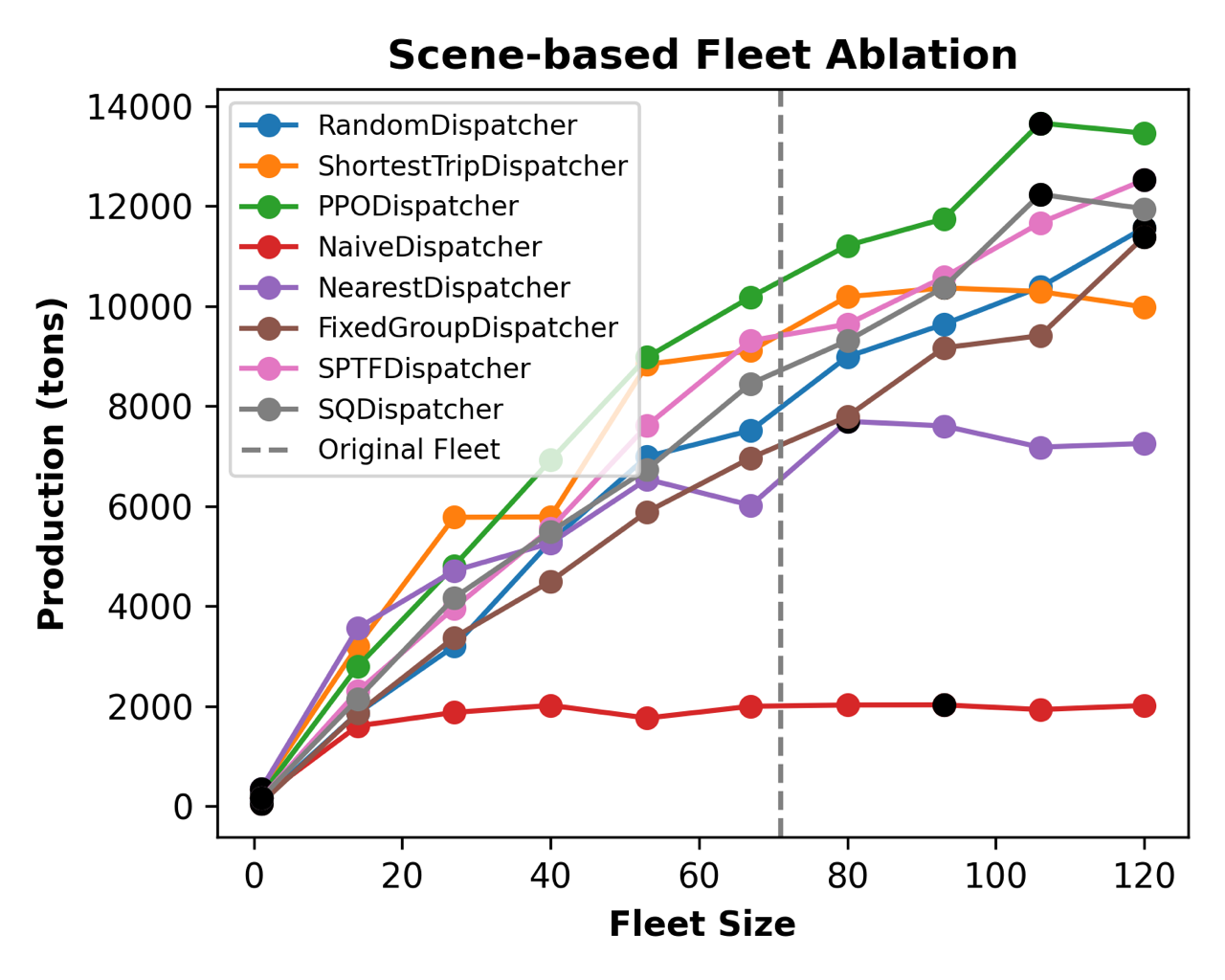} 
   \caption{Fleet Ablation Comparison}
   \label{fig:fleet_ablation}
\end{figure}

We used Optuna~\cite{akiba2019optuna} for 500 trials of hyperparameter search to obtain optimal hyperparameters, such as learning rate, clip ratio, and entropy coefficient, aiming to maximize the final production tonnage. Based on the optimal hyperparameters, we trained for 30 million steps each using Sparse Reward, Dense Reward, guided Dense Reward, and guided Sparse Reward, as shown in Fig. \ref{fig:curve}.  Simultaneously, we also horizontally compared the production performance of the reinforcement learning algorithm (guided dense reward, 30M steps) under different fleet sizes, as shown in Fig. \ref{fig:fleet_ablation}.

\subsection{Experimental Results}
\begin{figure}[t]
  \centering
   \includegraphics[width=0.5\textwidth]{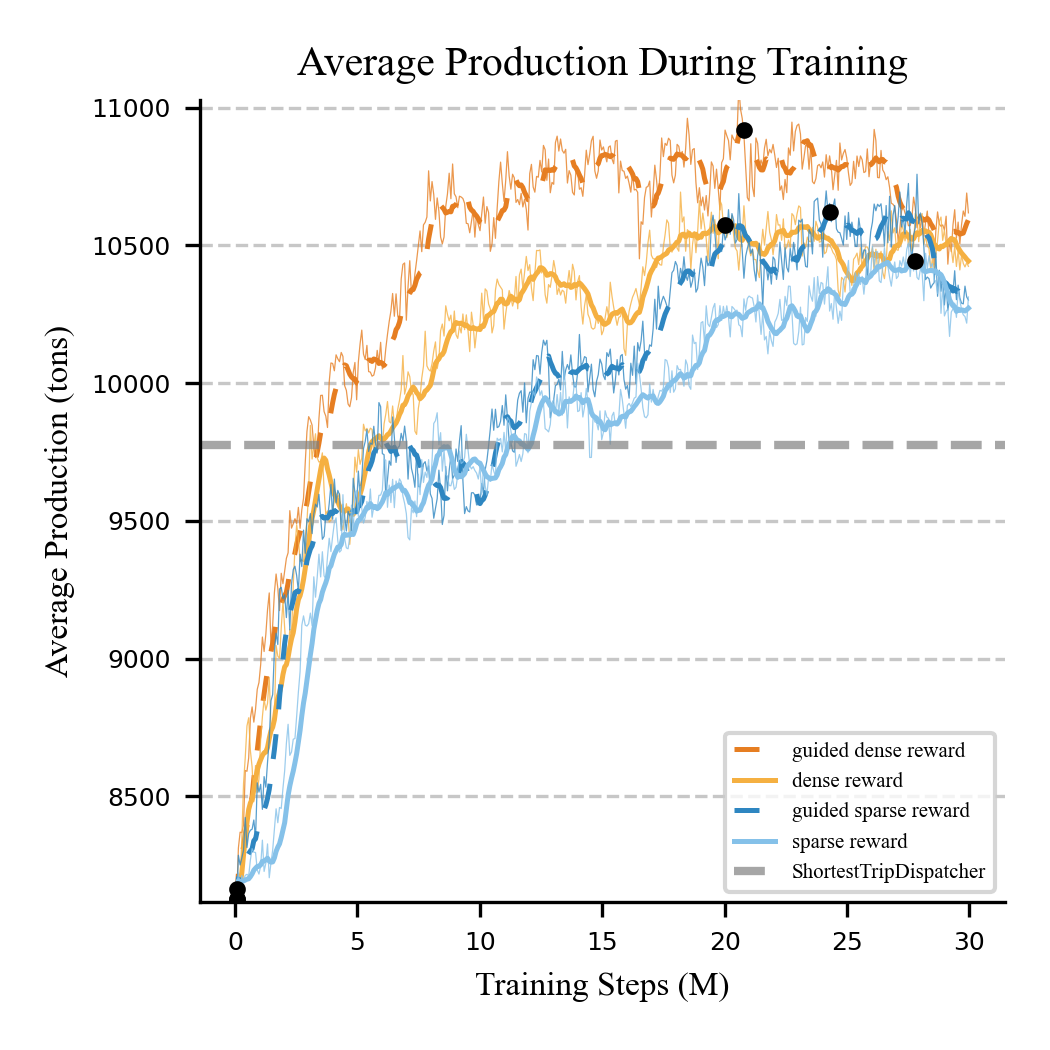} 
   \caption{Production Performance During Training}
   \label{fig:curve}
\end{figure}

To evaluate the effectiveness of adaptive direct policy guidance, we plotted the production performance of the Agent during training as shown in Fig. \ref{fig:curve}, comparing the performance of sparse and dense rewards with direct policy guidance, respectively.  We also horizontally compared the trained Agent with traditional strategy algorithms and made comparisons of their performance under different fleet sizes, which are presented in Table \ref{tab:performance_metrics} and Fig. \ref{fig:fleet_ablation} respectively.





\begin{table}[htbp]
\centering
\resizebox{\linewidth}{!}{
\begin{tabular}{@{}lcccc@{}}
\toprule
\textbf{Name} & \textbf{Produced} & \textbf{Matching} & \textbf{Total Wait} & \textbf{Jam} \\
& \textbf{Tons} & \textbf{Factor} & \textbf{Time} & \textbf{Ratio} \\
\midrule
FixedGroupDispatcher & 7401.74  & 0.45 & 1037.95 & 0.66 \\
NaiveDispatcher & 1941.14  & 0.46 & 10426.48 & 0.42 \\
NearestDispatcher & 7265.05  & 0.38 & 3720.43 & 0.38 \\
PPODispatcher & 11058.66 & 0.62 & 888.48 & 0.28 \\
RandomDispatcher & 8218.74  & 0.49 & 834.35 & 0.46 \\
ShortestTripDispatcher & 8703.40 & 0.58 & 2493.32 & 0.34 \\
SQDispatcher & 8690.11  & 0.51 & 840.36 & 0.41 \\
SPTFDispatcher & 9018.17  & 0.54 & 577.87 & 0.29 \\
\bottomrule
\end{tabular}
}
\caption{Dispatcher performance result.}
\label{tab:performance_metrics}
\end{table}

Fig.~\ref{fig:curve} shows that the policy guided by the teacher policy can usually achieve about 5\% performance improvement under the same computing power, while the reinforcement learning policy is about 10\% better than the original teacher policy overall. Under sparse rewards, the policy encounters difficulties in starting up in the early stages of training, and as the number of training steps increases, performance progresses slowly. The sparse policy guided by the direct policy, however, achieves better training performance earlier. Artificially designed dense rewards can also improve the training convergence speed and final performance, and teacher policy guidance has the most significant effect on the convergence speed under dense rewards, and has the best output performance. This is because in the early stage, the teacher-guided policy acts as a heuristic method for policy search and dense reward function, which helps to reduce the exploration space, thus making training more efficient.

Fig.~\ref{fig:fleet_ablation} uses gray dashed lines to mark the actual fleet size (71 heterogeneous mining trucks) for training and conducts experiments between 1 and 120. The experiments show that reinforcement learning algorithms can maintain good performance through generalization in the dispatching tasks of medium and large fleets of mining trucks. However, in smaller fleet sizes, the performance of reinforcement learning algorithms is not as good as traditional rule-based methods. This may be because random events encountered by mining trucks in small-scale fleets are less frequently triggered, resulting in the behavior of the environment overflowing the fitting interval of reinforcement learning training.

In Table~\ref{tab:performance_metrics}, the Jam Ratio indicates the proportion of Truck Trips encountering traffic jams. The indicators show that the reinforcement learning-based algorithm achieves better output performance in the new road network configuration. The traditional fixed-group strategy, however, performs poorly due to its lack of flexibility to random events (such as JamEvent) and its greedy selection of the nearest unloading point. A Match Factor of 0.62 indicates that the fleet size can still be increased to achieve better performance in the current scenario.

\section{CONCLUSIONS}
Mine fleet dispatching algorithms are crucial for efficient mine production. However, most academic work is concentrated in the traditional field of operations research optimization. Reinforcement learning algorithms, starting from the sequential decision-making process, are more adaptable to complex scenarios and have a higher development ceiling. Current reinforcement learning dispatching algorithms are typically solved in a value-based manner and require significant manual effort in reward function design and reward shaping. There is a lack of training methods that are robust to reward function design and can overcome cold start difficulties even under sparse rewards. This study starts from the PPO (Policy Based) reinforcement learning method, combines it with the practical application characteristics of mine fleet dispatching algorithms, and explores it using the OpenMines mine fleet dispatching simulation environment. It proposes a novel reinforcement learning dispatching algorithm based on direct policy guidance, which performs curriculum-based adaptive policy teaching by constructing new regularization terms, demonstrating good generalizability and performance. This curriculum-inspired adaptive direct policy guidance method is not only applicable to PPO-like algorithms but also to DQN-like value-based methods. Based on this research, future work can explore more complex model architectures, such as neural network architectures based on fine-tuning end-side large language models, thereby introducing the ability of RL Agents to follow instructions described in language.

\addtolength{\textheight}{-12cm}   






\bibliographystyle{./IEEEtran} 
\bibliography{./IEEEabrv.bib,./IEEEexample.bib}

\end{document}